# Leveraging Deep Learning Approaches for Deepfake Detection: A Review

ANIRUDDHA TIWARI

Minnesota State University, Mankato

DR. RUSHIT DAVE

Minnesota State University, Mankato

DR. MOUNIKA VANAMALA

University of Wisconsin-Eau Claire

*Abstract*— Conspicuous progression in the field of machine learning (ML) and deep learning (DL) have led the jump of highly realistic fake media, these media oftentimes referred as deepfakes. Deepfakes are fabricated media which are generated by sophisticated AI that are at times very difficult to set apart from the real media. So far, this media can be uploaded to the various social media platforms, hence advertising it to the world got easy, calling for an efficacious countermeasure. Thus, one of the optimistic counter steps against deepfake would be deepfake detection. To undertake this threat, researchers in the past have created models to detect deepfakes based on ML/DL techniques like Convolutional Neural Networks (CNN). This paper aims to explore different methodologies with an intention to achieve a cost-effective model with a higher accuracy with different types of the datasets, which is to address the generalizability of the dataset.

CCS CONCEPTS • Computing Methodologies ~ Machine Learning • Security and Privacy ~ Human and Societal Aspects of Security and Privacy • Computing Methodologies ~ Artificial Intelligence ~ Computer Vision ~ Computer Vision Problems ~ Object Detection

**Additional Keywords and Phrases:** Deepfake, Machine Learning, Deep Learning, Fake Image Detection, convolutional neural network, Long Short-Term Memory, Deep Neural Network.

## 1 INTRODUCTION

In the past few years, people have faced emerging issues with AI-made face-swapped videos or images. Specifically, machine learning fabricates images/videos in such a way that it is usually difficult to tell them apart from the real ones, and for these reasons they are called deepfakes. Deepfakes are primarily synthetic media in which a person's face in the image or video has been replaced by the likeliness of some other person. For instance, consider a video in which a reporter is reading the news; now replace his face with Person B. Now, the result is that Person B is reading the news in the video. Various techniques have already been studied by researchers to distinguish between the fake and real images, which include machine learning techniques like

Support vector machine (SVM) [1-3] and deep learning techniques like Convolutional Neural Networks (CNN) [4-6]. There are two most common ways of creating deepfakes, which are Generative adversarial networks (GANs) and Auto-encoders, which have the ability to misguide people by developing a high-dimensional pseudo image/video and making the audience believe that a well-known person is speaking/acting, whereas, in reality, he is not.

GANs are hard to train and difficult to use as a computational technique because they consist of a set of two neural networks, called generators and discriminators. This technique of deepfake generation believes that the generator should fool the discriminator; this makes the fakes more like real video and is significantly harder for the human eye to distinguish. Similarly, Autoencoders are another well-known deep learning technique; they have an encoder and a decoder function that carries some shared weights. Encoders and decoders can be utilized to take compressed details of an image to outsmart the existing compressing standards [7]. There are three different ways of using face-swapping techniques, which are lip syncing (in which the lip region in the target image is replaced by another person's lip region and can make someone speak what they originally never spoke); face-swapping (in which the person's face in the target image is replaced by the person's image in source image) [8]; and the most dangerous one, head-puppetry (in which target person's face is animated through the person sitting in front of the camera). Figure. 1 is a very good example of head puppetry [9].

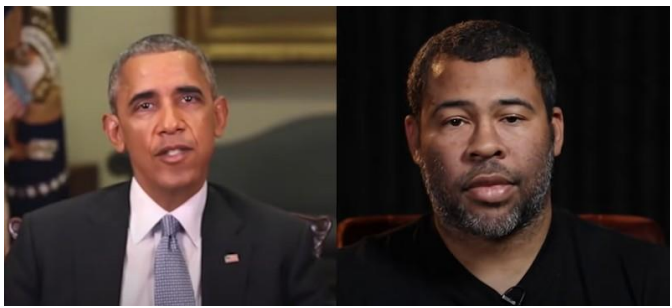

Figure 1. Example of head puppetry deepfake of President Barak Obama, from [9]

Deepfakes could be circulated over the globe with false news and people tend to believe and find it credible, which is an emerging threat to society. Detection of the deepfakes in the early years of the issue was possible even with the naked eye since you could see a color mismatch, the low resolution of synthesized faces, and sometimes temporal flickering [7], but since the current methodologies have transmogrified, detecting it with our eyes is not even possible, it is of that high quality [11]. Deepfake has been a source of misinformation and malevolence since it was introduced, causing a threat to society and political agendas [12]. At the depth of deepfake is deception—which is knowingly, technically, or intentionally, causing someone to believe something which is not true, typically to address personal, political agendas[13].The repercussions of deception by deepfakes are far more perceptible compared to that of verbal deception, it not only changes the oral content, but it also fabricates the visual properties of the videos/images on how the message was delivered [13].

Deepfakes produce tremendous threats for the future where fake news can be seen everywhere and can cause social impact, this unrestrained technological change deforms the truth, many times it could be intended for fun purposes, but often it is not. Therefore, we think there is a necessity for an effective, efficient detection techniques/algorithm which could stop the creation and usage of fake videos and evidently ML/DL techniques are good assets to recognize the human activity pattern to fight the threat [14-16]. In this work, we will provide an overview of the current state-of-the-art techniques of deepfake detection models and discuss the limitation and areas for future work. The later part of the paper is structured as follows: Section 2 will talk about the literature review of the deepfake methods opted in the past; Section 3 provides the discussion and analysis of the work based on the dataset, methods, etc., and addresses the research question; Section 4 reports the limitations of the work done and states the future development scope; Section 5 summarizes the paper followed by the references.

## 2 RELATED WORK

Over the past few years, people have been facing an emerging issue of AI synthesized face swapping videos/images which are widely known as deepfakes. This kind of fake video/image could be dangerous to society and cause threats to privacy, and fraudulence and can leave us vulnerable [17-19]. Looking towards the exponential growth of the deepfakes on social media platforms, we need a countermeasure to fight this threat. Researchers have been trying to detect deepfakes for a long time, but from the last few years we could evidently see some progress in the detection techniques. Deepfake detection is a binary classification problem [12][13] that justifies the authenticity of the media; hence it requires an ample amount of data which is the combination of real and fake videos that will be used to train the model. There are quite a few publicly available deepfake datasets which are Celeb-Deepfake (Celeb-DF) [7], FaceForensics++ [13], FFIW10K [13], Deepfake Detection Challenge (DFDC) [21], WildDeepfake [22], and more.

### 2.1 Convolutional Neural Network

Various methods have been investigated to classify deepfakes which rely on the facial discrepancies and artifacts in the video frame, inconsistency in the temporal correlations [23], discrepancies in the audio and visual modalities [24], or the spatial information from the images [25-26]. Researchers in [7] with the thinking of finding an algorithm or technology that can find out whether a photo was manipulated by face-swapping technology or not, present the DenseNet169 model with face-warping artifacts as indicators. For the negative data, they used photos of people and added some noise to them. Gaussian blur, Exponential blur, and Rayleigh blur were applied. In testing, they do not usually get Gaussian noise after affining the source picture, so they decided to try some other variants. They extracted faces from the images using the dlib package, after which they used random affine transformation on randomly resized pictures, followed by adding random specific blur, and finally, the images are resized, and real images are formed. For the evaluation purpose of the model Celeb-DF dataset was used, the dataset contained a total of 1000 high-quality videos whose certain parts were synthesized with no artifacts of the original face and small moving parts which makes it challenging for the model to detect.

Fabricated videos can be generated through Autoencoders and GANs, Authors of [27] focuses on creating deepfakes along with detection. Firstly, to create the deepfakes they utilize two autoencoders, the first autoencoder learns the feature of the source, whereas the second autoencoder learns the feature of the target image. Post feature extraction, the target image is reconstructed by the source image's decoder, this generates a target media that has features of the source image, and then the author proposes the use of DFDNet, a deep learning image enhancement method to improve the quality of the created deepfake, which will be an input to the detection model. For the detection model authors suggested the use of a deep neural network with a smaller number of layers called MesoNet, this methodology pays attention to the minute-to-minute details of the compressed frames, it starts with four layers of successive Convolutions and pooling which is followed by a dense network of the hidden layer. The model was trained with a dataset of more than 5000 images which was classified as real and fake, like the dataset used in [7]. As a result, the model was able to detect deepfake with a confidence interval of 80%.

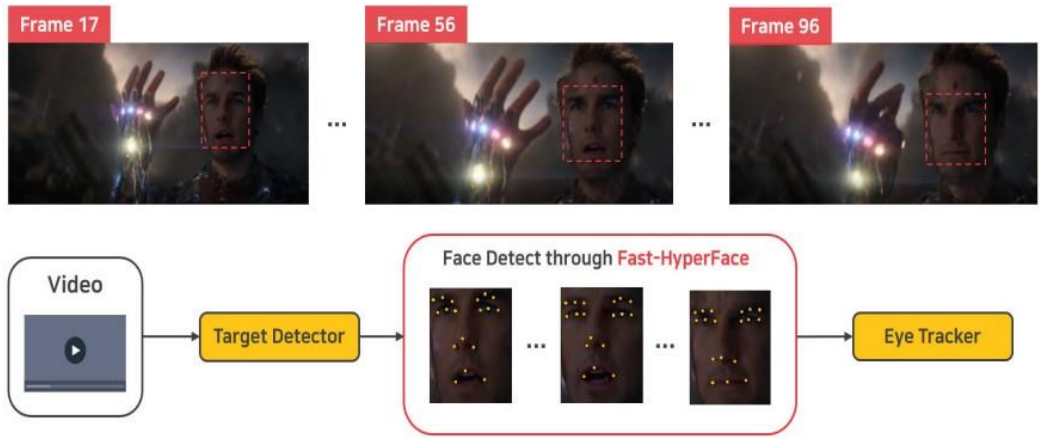

Figure 2. Visualization of DeepVision' s eye tracker. Which can measure the eye blinking, period through the Eye-Aspect-Ratio algorithm [11]

Investigation of [28] has identified that faces in many deepfake videos do not blink their eyes. However, many new cases have adjusted the discriminator to alter the eye blinking pattern has come into play to outwit the detection technique. To address that issue, the authors of [11] proposed to detect deepfakes through an algorithm called DeepVision. DeepVision analyzes significant changes in the eye blinking pattern which is a spontaneous action that depends on the overall physical condition of a human, age, and gender. Since eye blinking is a unique action that happens iteratively and occurs unconsciously, provides an alternative way of classifying deepfakes. This study implements an algorithm that analyzes and observes various behavioral and cognitive indicators that affect eye blinking patterns, which turns out to be a useful method to verify the integrity of the media. Blinking frequency fluctuates based on what activity a person is performing. While reading out loud or performing an act the frequency increases whereas when silently reading a book frequency decreases [11]. Authors have observed quite a few deepfake videos for the blinking pattern, and all of them showed an unnatural visual effect and had less than 5 blinks per minute, which is significantly less than the average eye blink rate. The frequency of the average eye blink between males and females differs. The architecture of the proposed model is divided into four parts, Input Data is the first stage which has video and data (gender, age, activity, time) as input. The second stage is algorithm one: target detector, using the fast-hyperface it detects the face in the frame, marks an outer line and sends it to the third stage algorithm 2: Eye tracker, which detects the eyes in the frame and calculates the frequency, and forwards it to the fourth stage algorithm 3: A method of comparative analysis which performs the integrity verification through the comparing and analyzing information of the blinking eyes as it could be seen in Figure 2. As a result, the proposed DeepVision model showcased an accuracy of 87.5% in verifying the integrity of the deepfakes.

When the feature is extracted from the images, usually it extracts relatively single artifacts, which often times leads to the low performance of the model, [29] suggesting the use of an attention mechanism fused with the detection model, which will extract the global and local feature of the face, hence improving the model accuracy. They tried to infuse this mechanism with XceptionNet, ResNet-18, and ResNet-50 and replace the ReLU activation function with Swish activation function as it tends to perform better with the deep neural network structure [30]. The FaceForensics++ dataset was used for experimental purposes, and the model showcased significant results in classification and Figure 3 shows the overall architecture proposed by the authors based on the ResNet50 model.

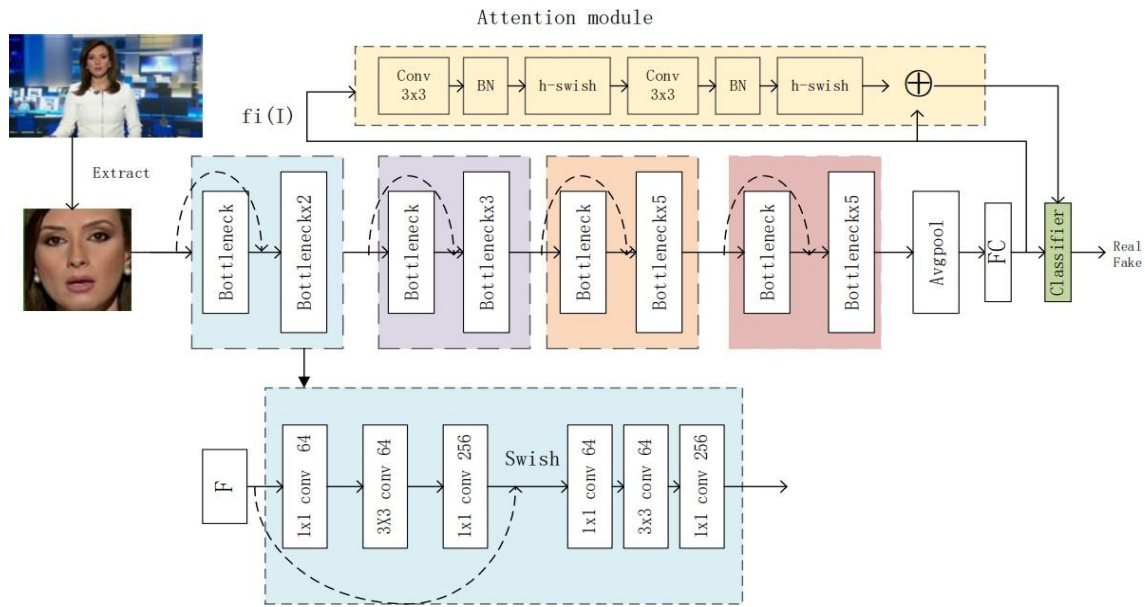

Figure 3. ResNet-50 model infused with the attention mechanism [29].

The application of MesoNet has been seen in many studies [27] [31], which focuses on building a structure with a low number of layers that could investigate the mesoscopic features of the images based on the image noise. However, their proposed structure could not be used in the compressed media as it degrades the background noise. The proposed solution utilized two networks namely Meso-4 and MespoInception-4 which showcased an average accuracy of 98% on the deepfake dataset and 95% on the FaceForensics++ data set. Another novel approach was introduced by the authors of [32], they proposed to classify deepfake based on a hybrid architecture of CNN and VGG16 and outsmarted the state-of-the-art techniques, and the Deepfake predictor (DFP) acquired an accuracy and precision of 94%. A study by [33] proposes a method that was an affinity of two methods which distinguishes the fakes by the inconsistent head possess and classifying based on the ResNet-50. In addition to Resnet, the authors chose to use a classifying algorithm named super-resolution algorithm, with an aim to increase the prediction accuracy of the low-resolution video. As a result, the proposed model had an accuracy of 94.4% on the UADFV dataset.

Another impressive work is done by the researchers of [24] who propose a deep neural network (DNN) based framework and an algorithm to classify deepfakes with a vision to address three challenges that lie in the very origin of the deepfakes which are: 1) effectively detect deepfakes; 2) Model applicable to compressed videos; and 3) creating a lighter version of the model. FaceForensics++ dataset by google with different compression levels was used to train and evaluate the proposed model. To address the mentioned challenges authors suggested creating an algorithm with low complexity, with a network consisting of two modules, (1) a CNN for frame feature extraction, and (2) a classifier network for the detection of fake videos. Three different CNN modules which are ResNet50, InceptionV3, and XceptionNet were used initially to get the best feature extractor for the compressed videos, these models were implemented over two compression levels which are c=23 and c=40. XceptionNet was chosen as the feature extractor for the suggested model as it shows better accuracy than all other models.

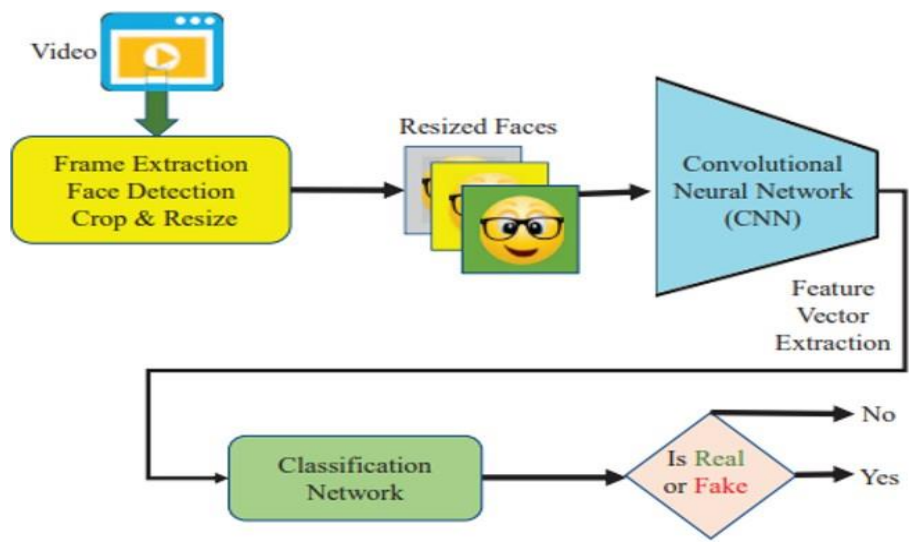

Figure 4. Detailed presentation of the proposed architecture [24].

**2.2 Long Shot-Term Memory**

Recent progression [34-35] has comprehensively changed the way of tempering images and videos. Researchers of [36] aim to detect video deepfakes by building Deep Neural Networks like Long short-term memory (LSTM) and InceptionResNetV2, which is a combination of ResNet and InceptionNet with 164 layers for detecting the object and extracting the features out of an image. Studies have identified that LSTM solves the vanishing gradient problems in recurrent neural networks and is architectured in a way that they learn long-term dependencies on the input data and sequentially process them [36]. The authors used pre-trained InceptionResNetV2 CNN to extract features, as using the pre-trained model helps to reduce the training and size difficulties. The extracted features from InceptionResNetV2 are used to train a 2048 LSTM layer that analyzes whether or not the video has been manipulated at different times because LSTM processes the input sequentially and compares the frames at different timings to see the matches in the frames and classifies them accordingly. Additionally, the model is also able to detect short clips/portions of the video if that has been fabricated. However, in the work of [37], authors focused on investigating the intra-frame modeling and inter-frame features to validate the authenticity of the video. To extract the temporal features, the authors leveraged an optical flow-based feature which is then fed to the model which is a combination of CNN and Recurrent neural network (RNN) architecture, the model had an accuracy of 92% with the FaceForensics++ dataset and Figure 5 showcase the proposed workflow.

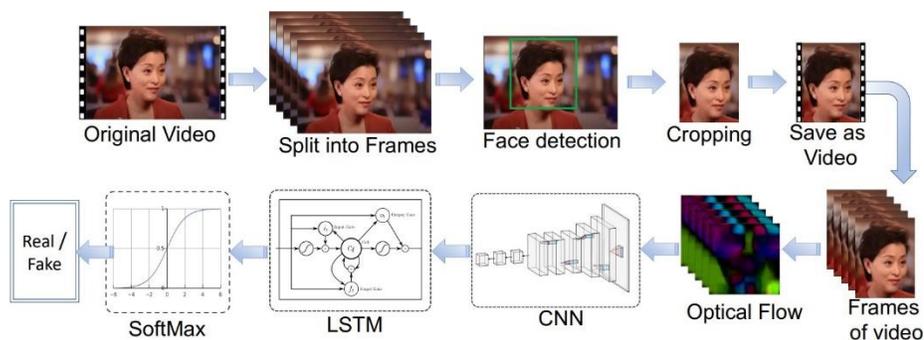

Figure 5. Proposed workflow to detect deepfake [37].

Many studies that we see either look for temporal granularity or visual artifacts, whereas researchers in [28] found a way to distinguish deepfakes based on the human eye blinking pattern like the authors of DeepVision [11]. To find the difference between open and closed eye states, the authors propose the use of a deep neural network model which combines CNN with RNN and is known as a Long-Term Recurrent Network. Figure 6 shows the proposed architecture of the LRCN model. Research work of [38] proposes a layered approach where they used a facial recognition network to detect the target, then used CNN to extract the features out of the target, which uses gaussian blur and gaussian noise which could disregard the high-frequency noise and sounds and then the features are passed on to the LSTM layer where they made a sequence of temporal features in the manipulated faces between the frames. In addition to that the authors used Recycle-GAN, the advantage of using Recycle-GAN is that while it is evaluating, it sends the results back to start and fed the model again so that it could improve its mistakes if made any and can manipulate the parameters accordingly.

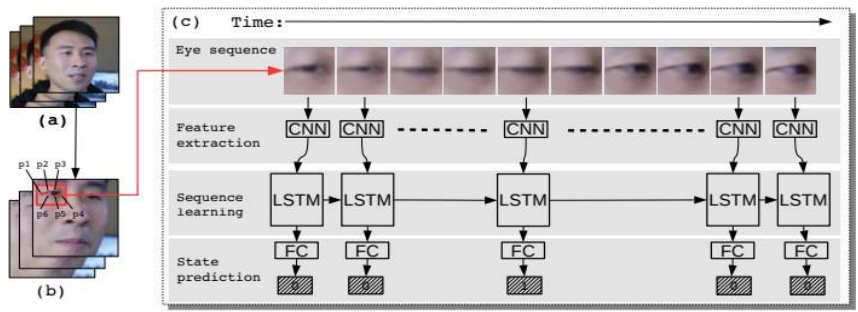

Figure 6. Overview of the LRCN method [28]

## 3 DISCUSSION AND ANALYSIS

In this paper, we have discussed many existing methods and techniques to create and detect deepfakes. With every passing day the technology is improving and so does the method to create more highly realistic deepfakes. Thus, making the detection work difficult. Deepfakes are often created as an act of revenge, to address political agendas, or for some fraudulent activities. To continue the study and development of the new models and architectures, it is highly important to have a dataset that contains highly realistic images and videos from which the frames could be extracted [39]. Our study depicts that researchers have identified the use of various deep learning-based algorithm to classify deepfakes, this algorithm includes CNN, DNN, RNN, LSTM, and further CNN could be infused with some other techniques to manipulate few parameters that result in identifying the deepfakes from the real one. Few of the authors tried to find discrepancies in the visual artifacts, few focused on finding the changes in the human behaviors or traits that a human could only possess. Some of the studies perform best in identifying the changes in the eye blinking pattern [11]. These models were evaluated on different datasets named FaceForensics++, Celeb-DF, UADFV [40], WildDeepfakes, and DFDC. UADFV consisted of a total of 34.6K images among which 17.3k were forged and 17.3k were real. These datasets were created with different deepfake generation techniques whether it be GANs, Autoencoders, or any other image manipulation techniques.

The aforementioned datasets provide a good gauge to evaluate different models to each other, as the outcomes could be significantly dependent on the dataset. As mentioned earlier each specific model can detect deepfake that are either generated through GANs or Autoencoders. Therefore, not one algorithm could classify two different types of datasets. Looking from the high level, the studies that utilized the method to analyze the temporal features performed well, whereas there was little number of studies that focused on human behavior or patterns. However, the DeepVision model performed well and showcased an accuracy of about 87.5% in authenticating the identity. The result from the studies portrays the efficacy of the Deep learning model like CNN and Other neural networks like MesoNet, Inception Net, XceptionNet, etc. Overall, the models showcased a highly significant accuracy, through which we could say that there is a lot of improvement in the deepfakes detection techniques, and future developments could count on the exemplary work done by the authors to combat the threat from deepfakes around the globe. As previously mentioned, CNN based deepfake classifying model performs outstandingly only when tested and trained within the same type of dataset. However, once these models are tested over the other new datasets the performance of the models drops significantly. It would be better to say that these models are only fitted for the intra-dataset evaluation.

**Research Question**: "How can we efficiently build a deepfake detection model while keeping the transferability over various datasets and higher accuracy intact?"

## 4 LIMITATION AND FUTURE WORK

In these times the deepfakes are significantly increasing, thus as a consequence, detection of the fake media is another big task in this online world, or we can say in the world of digital media forensics. Existing models to classify the falsification of videos or images are dependent on a specific method of creation of deepfakes. One model only shows high accuracy when the deepfakes were generated through GANs, but here if this model is being tested on a dataset created through autoencoder, it will affect its accuracy negatively. The most common way of detecting fake videos is by analyzing facial artifacts and temporal features, but for the high-quality deepfakes, these models and methodologies show lower accuracy and models tend to not work effectively. Models have a tendency to not detect fake media if the facial structure is straight pointing towards the camera, since the picture cannot identify color changes or resolution changes in the image and similar happens for the head posture as well.

Table 1: Summary of CNN based work.

| Title | Methods | Results | Dataset | Contribution |
|---|---|---|---|---|
| Methods of deepfake detection based on machine learning [7] | DenseNet169 + Rayleigh Blur | Accuracy: 60% | Celeb-DF | Found the indicators that can distinguish whether face manipulation is applied on any media. |
| DeepVision: Deepfakes detection using eye blinking pattern [11] | DeepVision with integrity verification | Accuracy: 87.5% | Eye Blinking Prediction Dataset from Kaggle. | Invented new model that was able to set apart fake videos which has inconsistent eye blinking with significant accuracy. |
| Deepfakes creation and detection using deep learning [27] | MesoNet with DFDNet image enhancer | Accuracy: 80% | Online dataset containing 5000 images. | Application of image enhancer over the images increased the classification accuracy of the MesoNet. |
| Combining deep learning and super-resolution algorithms for deepfake detection [33] | ResNet-50 with super resolution pre-processing. | Accuracy: 94.4% | UADFV | Model with super resolution had good accuracy. However, could not perform better on the head pose estimator. |
| A novel machine learning based method for deepfake video detection in social media [24] | ResNet-50, InceptionV3, XceptionNet | Accuracy: 88%, 86%, 96% | FaceForensics++ | Trained the model with immediate compression, which resulted in significant increase in accuracy. |

If we talk about the model that classifies images based on the eye blinking pattern, would only focus on the eye region. However, if the same media have lip area forged this model would not be able to classify and no direction for the fast eye blinking is described [11] [28]. With the aforementioned limitations and gaps, we would want to investigate finding a way to efficiently and securely detect fake media's using ML/DL techniques with the development of a new model and taking the new features into account. Additionally, we would like to study more on how to generalize it with both image and video datasets created with either GANs or Autoencoders method. So, the models do not only perform on specific deepfake generating methodology. Another problem that could be considered in the future would be to think about the cost of building these models. Future research could also be directed in to find the optimal way of detecting discrepancies on the whole face regardless of eye region or the lip region or generally any region on the face. However, in-depth knowledge and study would be required to continue further research.

Table 2. Summary of work based on LSTM.

| Title | Method | Result | Dataset | Contribution |
|---|---|---|---|---|
| Deepfake Detection using InceptionResnetV2 and LSTM [36] | InceptionResNetV2 with LSTM | 20 Epochs: Accuracy 84.75% 40 Epoch: Accuracy 91.48% | Celeb-DF | Basis of their model is to look for the left-over traces which are not visible by naked eye and due to which they acquired significant accuracy. |
| A hybrid CNN-LSTM model for video deepfake detection by leveraging optical flow features [27] | CNN-Optical Flow-LSTM | Accuracy: 91.21% | FaceForensics++ | Integrated Optical flow features in the CNN model to gain higher accuracy. |
| In Ictu Oculi: Exposing AI Created Fake Videos by Detecting Eye Blinking [28] | CNN-VGG and LRCN | Accuracy: 96.2% | CEW Dataset | Introduced the LRCN (Long-Term Convolutional Network) to capture the eye blinking pattern. |

## 5 CONCLUSION

Creating deepfakes is growing as it is getting more convenient even for people with a minimal level of technological knowledge, thanks to the advanced technology researchers have developed, but detection of deepfakes remains more of a challenging issue that needs to be addressed sooner than later. Detecting deepfakes is crucial in today's digital age as it helps maintain the authenticity and integrity of information. With the increasing sophistication of deepfake technology, it is becoming easier to manipulate images, audio, and video to spread false information and misinformation. This not only undermines trust in legitimate sources of information, but it also has the potential to cause significant harm in areas such as politics, finance, and national security. With the rapid growth of social media platforms and our dependency on them, we can be certain of one thing which is the rise of more deepfakes (images/videos) and detecting deepfakes will continue to be an ongoing crucial process.

In this paper, we have discussed the various machine learning and deep learning-based methodologies that can be utilized to detect deepfakes, among all the techniques that we discussed we found CNN and CNN integrated techniques to be more accurate in distinguishing between real and fake media. Here, the researchers have used various kinds of dataset which includes the fake images that were created through different methods, and we also mentioned the WildDeepfake dataset that was created by [15] which consists of 1,180,099 images that were extracted from 707 videos. WildDeepfake could serve as a supporting dataset for new inventions and evaluation purposes. By developing and implementing robust deepfake detection methods, we can ensure that the information we consume is accurate and trustworthy and protects against the malicious use of deepfakes.

Obviously, there would be some limitations to our study and its findings. Throughout the paper, we have discussed and evaluated many articles, tools, and different techniques to understand how the deepfake is generated and the current situation of the detection techniques introduced by numerous other authors. However, there are still more scholarly articles available that need to be studied in depth, to find an absolute knowledge about this technology. A deep study on this topic would lead to the further research opportunities in the future.

**ACKNOWLEDGMENTS**

I would like to take this as an opportunity to appreciate my mentor Dr.Rushit Dave, who guided me throughout this study and for all his valuable support.